\documentclass{article}
\usepackage{ijcai19}

\usepackage{times}
\usepackage{soul}
\usepackage{url}
\usepackage[hidelinks]{hyperref}
\usepackage[utf8]{inputenc}
\usepackage[small]{caption}
\usepackage{graphicx}
\usepackage{amsmath}
\usepackage{booktabs}
\usepackage{multirow}
\usepackage{subfigure}
\usepackage{verbatim}
\usepackage{algorithm}
\usepackage{algorithmic}
\usepackage{makecell}
\usepackage{CJKutf8}
\usepackage{flushend}
\title{Neural News Recommendation with Attentive Multi-View Learning}

\author{Chuhan Wu$^1$\and
Fangzhao Wu$^2$\and
Mingxiao An$^3$\and
Jianqiang Huang$^4$\and
Yongfeng Huang$^1$\and
Xing Xie$^2$\\
\affiliations
$^1$Department of Electronic Engineering, Tsinghua University, Beijing 100084, China\\
$^2$Microsoft Research Asia, Beijing 100080, China\\
$^3$University of Science and Technology of China, Hefei 230026, China\\
$^4$Peking University, Beijing 100871, China\\
\emails
wuch15@mails.tsinghua.edu.cn,
\{fangzwu,xingx\}@microsoft.com,
anmx@mail.ustc.edu.cn,1701210864@pku.edu.cn,yfhuang@tsinghua.edu.cn}

\begin{document}

\maketitle

\begin{abstract}
Personalized news recommendation is very important for online news platforms to help users find interested news and improve user experience.
News and user representation learning is critical for news recommendation.
Existing news recommendation methods usually learn these representations based on single news information, e.g., title, which may be insufficient.
In this paper we propose a neural news recommendation approach which can learn informative representations of users and news by exploiting different kinds of news information.
The core of our approach is a news encoder and a user encoder.
In the news encoder we propose an attentive multi-view learning model to learn unified news representations from titles, bodies and topic categories by regarding them as different views of news.
In addition, we apply both word-level and view-level attention mechanism to news encoder to select important words and views for learning informative news representations.
In the user encoder we learn the representations of users based on their browsed news and apply attention mechanism to select informative news for user representation learning.
Extensive experiments on a real-world dataset show our approach can effectively improve the performance of news recommendation.

\end{abstract}

\section{Introduction}

Online news services such as Google News\footnote{https://news.google.com/} and MSN News\footnote{https://www.msn.com/en-us/news} which collect and aggregate news from many sources have become popular platforms for news reading~\cite{das2007google,lavie2010user}.
Tens of thousands of news are generated everyday, and it is impossible for users to read all these news articles due to time limit~\cite{phelan2011terms}.
Therefore, personalized news recommendation is very important for online news platforms to help users find their interested news and alleviate information overload~\cite{ijntema2010ontology,de2012chatter}.

\begin{figure}[!t]
  \centering
    \includegraphics[width=0.95\linewidth]{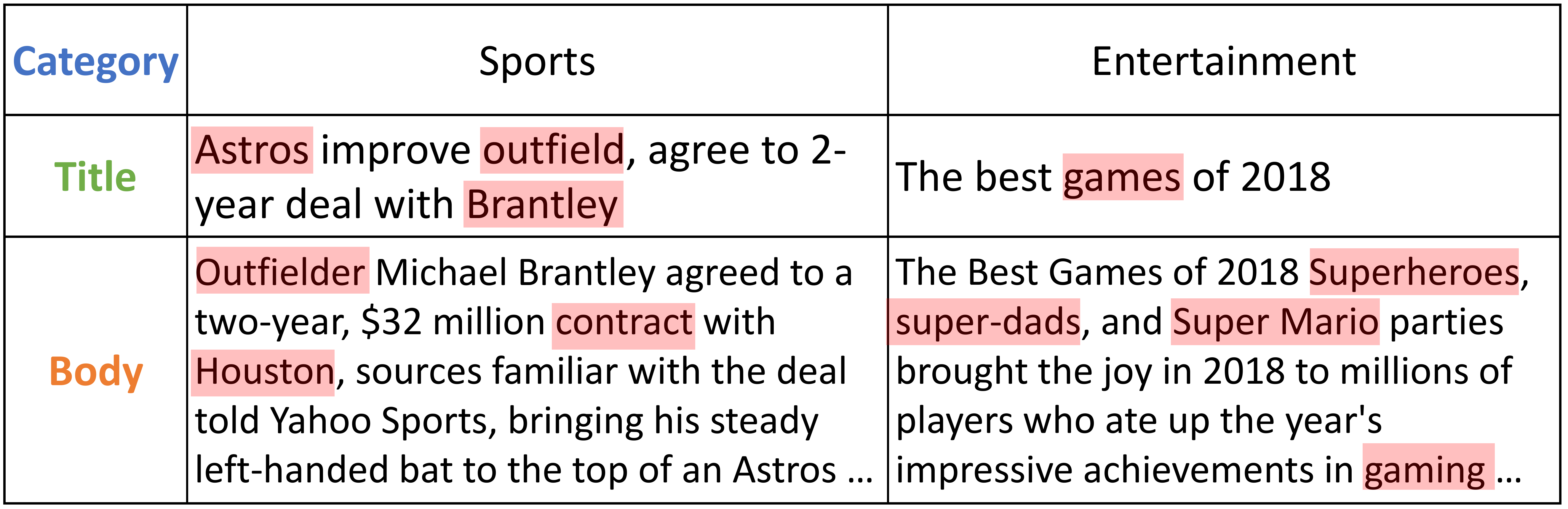}
  \caption{Two example news articles. Red bars represent important words in news titles and bodies.}
  \label{fig.example}
  \vspace{-0.15in}
\end{figure}

Many deep learning based news recommendation methods have been proposed~\cite{okura2017embedding,wang2018dkn}.
A core problem in these methods is how to learn representations of news and users.
Existing methods usually learn these representations based on single news information.
For example, Okura et al.~\shortcite{okura2017embedding} proposed to learn news representations from the body of news articles via auto-encoders, and then learn users representations from news representations by applying GRU to their browsed news.
Wang et al.~\shortcite{wang2018dkn} proposed to learn news representations via a knowledge-aware CNN from the titles of news articles, and then learn user representations from news representations based on the similarity between candidate news and each browsed news.
The single kind of news information may be insufficient for learning accurate news and user representations, and may limit the performance of news recommendation.

Our work is motivated by the following observations.
First, a news article usually contains different kinds of information, such as title, body and topic category, which are all useful for representing news.
For example, in Fig.~\ref{fig.example} the title of the first news indicates that it is about a deal between a baseball star and a team, and the body can provide more detailed information of this deal.
In addition, the topic category of this news is also informative, since if a user clicks this news and many other news articles with the same category, then we can infer she is very interested in sports.
Thus, incorporating different kinds of news information has the potential to improve news and user representations for news recommendation.
Second, different kinds of news information have different characteristics.
For example, the titles are usually very short and concise, while the bodies are much longer and more detailed. 
The topic categories are usually labels with very few words.
Thus, different kinds of news information should be processed differently.
Third, different news information may have different informativeness for different news.
For example, the title of the first news in Fig.~\ref{fig.example} is precise and important for representing this news, while the title of the second news is short, vague, and less informative.
Besides, different words in the same news may have different importance.
For example, in Fig.~\ref{fig.example}, ``outfileder'' is more important than ``familiar'' for representing the first news.
In addition, different news browsed by the same user may also have different informativeness for learning user representations.
For example, if a user browses a news of ``Super Bowl'' which is extremely popular and a news of ``International Mathematical Olympiad'' which is less popular, then the second news may be more informative for inferring her interests than the first one.

In this paper, we propose a neural news recommendation approach with attentive multi-view learning (NAML) to learn informative news and user representations by exploiting different types of news information. 
Our approach contains two core components, i.e., a news encoder and a user encoder.
In the news encoder, we propose an attentive multi-view learning framework to learn unified news representations from titles, bodies and categories by incorporating them as different views of news.
In addition, since different words and views may have different informativeness for news representation, we apply word-level and view-level attention networks to select important words and views for learning informative news representations.
In the user encoder, we learn the representations of users from the representations of their clicked news.
Since different news may also have different informativeness for user representations, we apply attention mechanism to news encoder to select informative news for user representation learning.
We conduct extensive experiments on a real-world dataset.
The results show our approach can effectively improve the performance of news recommendation.

\section{Related Work}\label{sec:RelatedWork}

News recommendation is an important task in natural language processing and data mining fields, and has wide applications~\cite{zheng2018drn,wu2019neuralnews}.
Learning accurate news and user representations is critical for news recommendation.
Many existing news recommendation methods rely on manual feature engineering for news and user representation learning~\cite{liu2010personalized,son2013location,bansal2015content}.
For example, Liu et al.~\shortcite{liu2010personalized} proposed to use topic categories and interest features generated by a Bayesian model as news representations.
Son et al.~\shortcite{son2013location} proposed an Explicit Localized Semantic Analysis (ELSA) model for location-based news recommendation.
They proposed to extract topic and location features from Wikipedia pages as news representations.
Lian et al.~\shortcite{lian2018towards} proposed a deep fusion model (DMF) to learn news representations from various handcrafted features such as title length and news categories.
However, these methods rely on manual feature engineering, which needs a large amount of domain knowledge and effort to craft.
In addition, these methods cannot capture contexts and orders of words in news, which are important for learning news and user representations.

In recent years, several deep learning based news recommendation methods are proposed~\cite{okura2017embedding,kumar2017word,khattar2018weave,wang2018dkn,wu2019npa}.
For example, Okura et al.~\shortcite{okura2017embedding} proposed to learn news representations from news bodies using denoising autoencoders. 
In addition, they proposed to learn user representations from their browsed news using GRU network.
Wang et al.~\shortcite{wang2018dkn} proposed to learn news representations from news titles via a knowledge-aware CNN network which can incorporate the information in knowledge graphs.
However, these methods can only exploit a single kind of news information, which may be insufficient for learning accurate representations of news and users.
Different from these methods, our approach can learn representations of news and users by incorporating different kinds of news information such as title, body and topic category via an attentive multi-view learning framework.
Extensive experiments on real-world dataset validate our approach can learn better news and user representations, and achieve better performance on news recommendation than existing methods.

\section{Our Approach}\label{sec:Model}
\begin{figure*}[!t]
  \centering
    \includegraphics[width=0.9\linewidth]{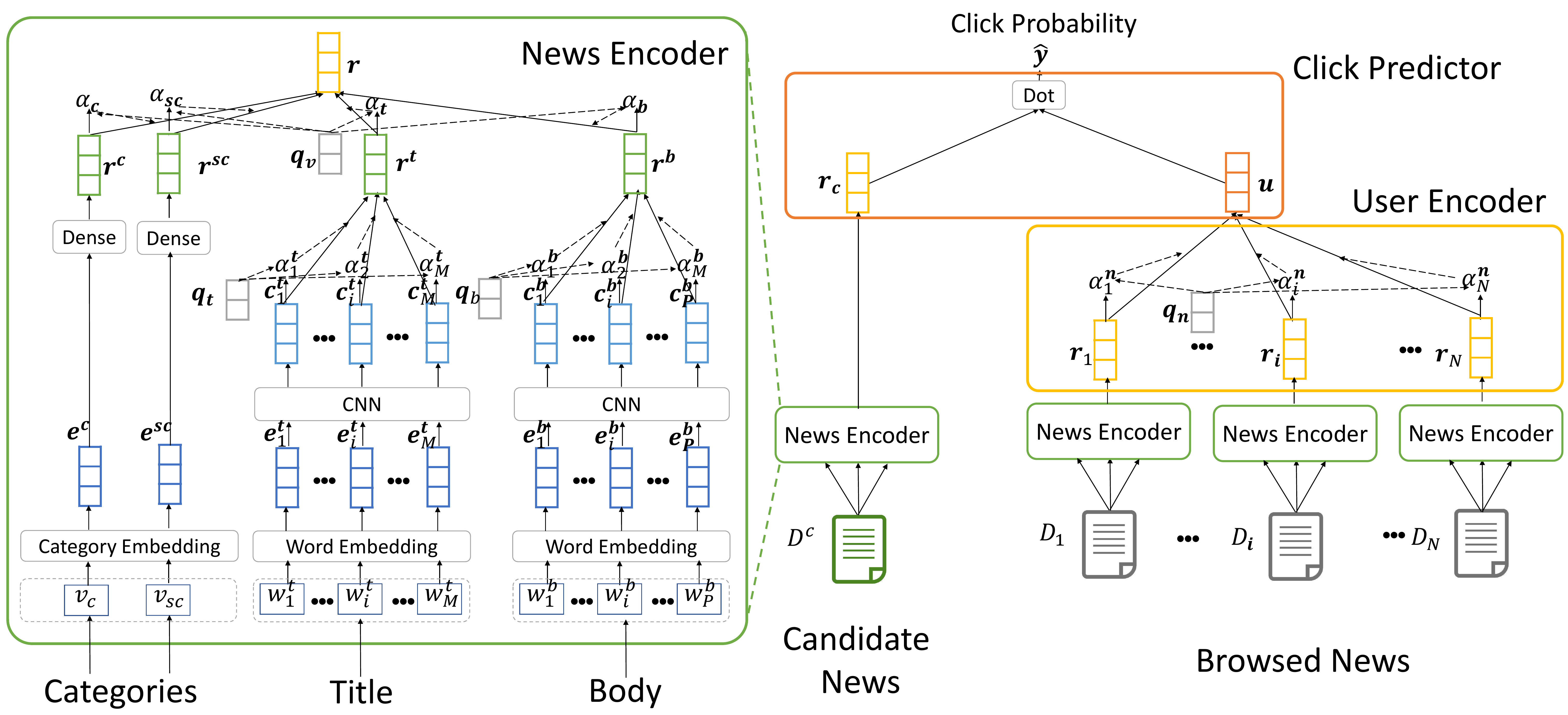}
  \caption{The framework of our \textit{NAML} approach for news recommendation.}
\vspace{-0.1in}
  \label{fig.model}
\end{figure*}

In this section, we introduce our \textit{NAML} approach for neural news recommendation.
There are three major modules in our approach, i.e., a \textit{news encoder} with attentive multi-view learning to learn representations of news, a \textit{user encoder} with attention mechanism to learn representations of users from their browsed news, and a \textit{click predictor} to predict the probability of a user browsing a candidate news article.
The architecture of our approach is shown in Fig.~\ref{fig.model}.

\subsection{News Encoder}

The \textit{news encoder} module is used to learn representations of news from different kinds of news information, such as titles, bodies and topic categories.
Since different kinds of news information have different characteristics, instead of simply merging them into a long text for news representation, we propose an attentive multi-view learning framework to learn unified news representations by regarding each kind of news information as a specific view of news. 
As shown in Fig.~\ref{fig.model}, there are four major components in \textit{news encoder}.

The first component is \textit{title encoder}, which is used to learn news representations from their titles.
There are three layers in the \textit{title encoder}.
The first layer is word embedding, which is used to convert a news title from a word sequence into a sequence of low-dimensional semantic vectors.
Denote the word sequence of a news title as $[w^t_1,w^t_2,...,w^t_M]$, where $M$ is the length of this title.
It is converted into a sequence of word vectors $[\mathbf{e}^t_1,\mathbf{e}^t_2,...,\mathbf{e}^t_M]$ via a word embedding look-up table $\mathbf{W}_e\in \mathcal{R}^{V\times D}$, where $V$ and $D$ are vocabulary size and word embedding dimension respectively.

The second layer is a convolutional neural network (CNN)~\cite{kim2014convolutional}.
Local contexts of words in news titles are important for learning their representations.
For example, in the news title ``Xbox One On Sale This Week'', the local contexts of ``One'' such as ``Xbox'' and ``On Sale'' are useful for understanding that it belongs to a game console name.
Thus, we use CNNs to learn contextual word representations by capturing their local contexts.
Denote the contextual representation of the $i$-th word as $\mathbf{c}^t_i$, which is calculated by:
\begin{equation}
\mathbf{c}^t_i = \mathrm{ReLU}(\mathbf{F}_t\times \mathbf{e}^t_{(i-K):(i+K)}+\mathbf{b}_t),
\end{equation} 
where $\mathbf{e}^t_{(i-K):(i+K)}$ is the concatenation of word embeddings from position $(i-K)$ to $(i+K)$.
$\mathbf{F}_t \in \mathcal{R}^{N_f\times (2K+1)D}$ and $\mathbf{b}_t \in \mathcal{R}^{N_f}$ are the kernel and bias parameters of the CNN filters, $N_f$ is the number of CNN filters and $2K+1$ is their window size.
ReLU~\cite{glorot2011deep} is the non-linear activation function.
The output of this layer is a sequence of contextual word representations, i.e., $[\mathbf{c}^t_1,\mathbf{c}^t_2,...,\mathbf{c}^t_M]$.

The third layer is a word-level attention network~\cite{wu2019neuraldemo}.
Different words in the same news title usually have different informativeness for learning news representations.
For example, in the news title ``Rockets End 2018 With A Win'' the word ``Rockets'' is more informative than ``With'' in representing this news.
Thus, recognizing the important words in different new titles has the potential to learn more informative news representations.
We propose to use a word-level attention network to select important words within the context of each news title.
Denote the attention weight of the $i_{th}$ word in a news title as $\alpha^t_i$, which is formulated as:
\begin{equation}
\begin{split}
a^t_i & =  \mathbf{q}_t^T\tanh(\mathbf{V}_t \times \mathbf{c}^t_i+\mathbf{v}_t),\\
\alpha^t_i & = \frac{\exp(a^t_i)}{\sum_{j=1}^M{\exp(a_j^t)}},
\end{split}
\end{equation}
where $\mathbf{V}_t$ and $\mathbf{v}_t$ are the projection parameters, $\mathbf{q}_t$ denotes the attention query vector.

The final representation of a news title is the summation of the contextual representations of its words weighted by their attention weights, i.e., 
$\mathbf{r}^t = \sum_{j=1}^M{\alpha^t_j  \mathbf{c}^t_j}$.

The second component in \textit{news encoder} module is \textit{body encoder}, which is used to learn news representations from their bodies.
Similar with the \textit{title encoder}, there are also three layers in the \textit{body encoder}.

The first layer is word embedding,
which is shared with the \textit{title encoder}.
Denote the word sequence of a news body as $[w^b_1,w^b_2,...,w^b_P]$, where $P$ is the length of the body.
Through the word embedding layer, it is transformed into a sequence of word vectors $[\mathbf{e}^b_1,\mathbf{e}^b_2,...,\mathbf{e}^b_P]$.
The second layer is a CNN network, which takes the word vector sequence as input, and learns the contextual word representations $[\mathbf{c}^b_1,\mathbf{c}^b_2,...,\mathbf{c}^b_P]$ by capturing the local contexts.
The third layer is an attention network.
Different words in the same news body may have different informativeness for news representations.
For example, in Fig.~\ref{fig.example} the word ``outfielder'' is more informative than ``familiar'' in representing this news.
Thus, we apply an attention network to the \textit{body encoder} to select important words in the context of each news body for learning more informative news representations.
Denote the attention weight of the $i_{th}$ word in a news body as $\alpha^b_i$, which is calculated by:
\begin{equation}
\begin{split}
a^b_i & =  \mathbf{q}_b^T\tanh(\mathbf{V}_b \times \mathbf{c}^b_i+\mathbf{v}_b),\\
\alpha^b_i & = \frac{\exp(a^b_i)}{\sum_{j=1}^P{\exp(a^b_j)}},
\end{split}
\end{equation}
where $\mathbf{V}_b$ and $\mathbf{v}_b$ are projection parameters, and $\mathbf{q}_b$ is the attention query vector.
The final representation of a news body is the summation of the contextual word representations weighted by their attention weights, i.e., 
$\mathbf{r}^b = \sum_{j=1}^P{\alpha^b_j  \mathbf{c}^b_j}$.

The third component in the \textit{news encoder} module is \textit{category encoder}, which is used to learn news representations from their categories.
On many online news platforms such as MSN News, news articles are usually labeled with topic categories (e.g., ``sports'' and ``entertainment'') and subcategories (e.g., ``basketball\_nba'' and ``gaming'') for targeting user interests which contain important information of these news.
For example, if a user clicked many news articles with the category of ``sports'', then we can infer that this user may be interested in sports news.
Thus, we propose to incorporate both the category and subcategory information for news representation learning.
The inputs of the \textit{category encoder} are the ID of the category $v_c$ and the ID of the subcategory $v_{sc}$.
There are two layers in the \textit{category encoder}.
The first layer is a category ID embedding layer.
It converts the discrete IDs of categories and subcategories into low-dimensional dense representations (denoted as $\mathbf{e}^c$ and $\mathbf{e}^{sc}$ respectively).
The second layer is a dense layer.
It is used to learn the hidden category representations by transforming the category embeddings as:
\begin{equation}
\begin{split}
\mathbf{r}^c & = \mathrm{ReLU}(\mathbf{V}_c \times \mathbf{e}^c+\mathbf{v}_c),\\
\mathbf{r}^{sc} & = \mathrm{ReLU}(\mathbf{V}_s \times \mathbf{e}^{sc}+\mathbf{v}_s),
\end{split}
\end{equation}
where $\mathbf{V}_c$, $\mathbf{v}_c$, $\mathbf{V}_s$ and $\mathbf{v}_s$ are parameters in dense layers.

The fourth component in the \textit{news encoder} module is \textit{attentive pooling}.
Different kinds of news information may have different informativeness for learning the representations of different news.
For example, in Fig.~\ref{fig.example}, the title of the first news is precise and contains rich information.
Thus, it should have a high weight in representing this news.
However, the title of the second news is short and vague.
Thus, the body and the topic category should have higher weights than the title for representing this news.
Motivated by these observations, we propose a view-level attention network to model the informativeness of different kinds of news information for learning news representations.
Denote the attention weights of title, body, category and subcategory as $\alpha_t$, $\alpha_b$, $\alpha_c$ and $\alpha_{sc}$ respectively.
The attention weight of the title view is calculated by:
\begin{equation}
\begin{split}
a_t & =  \mathbf{q}_v^T\tanh(\mathbf{U}_v \times \mathbf{r}^t+\mathbf{u}_v),\\
\alpha_t & = \frac{\exp(a_t)}{\exp(a_t)+\exp(a_b)+\exp(a_c)+\exp(a_{sc})},
\end{split}
\end{equation}
where $\mathbf{U}_v$ and $\mathbf{u}_v$ are projection parameters, $\mathbf{q}_v$ is the attention query vector.
The attention weights of other news information such as body, category and subcategory can be computed in a similar way.

The final unified news representations learned by the \textit{news encoder} module is the summation of the news representations from different views weighted by their attention weights:
\begin{equation}
\mathbf{r}= \alpha_c\mathbf{r}^c+\alpha_{sc}\mathbf{r}^{sc}+\alpha_t\mathbf{r}^t+\alpha_b\mathbf{r}^b.
\end{equation}

In our \textit{NAML} approach the \textit{news encoder} module is used to learn the representations of both historical news browsed by users and the candidate news to be recommended.

\subsection{User Encoder}

The \textit{user encoder} module in our approach is used to learn the representations of users from the representations of their browsed news, as shown in Fig.~\ref{fig.model}.
Different news browsed by the same user have different informativeness for representing this user.
For example, the news ``10 best NBA moments'' is very informative for modeling user preferences since it is usually browsed by users who are interested in sports, but the news ``The Weather Next Week'' is less informative since it is clicked by massive users.
Therefore, in \textit{user encoder} module we apply a news attention network to learn more informative user representations by selecting important news.
We denote the attention weight of the $i_{th}$ news browsed by a user as $\alpha^n_i$, which is calculated as follows:
\begin{equation}
\begin{split}
a^n_i &=  \mathbf{q}_n^{T}\tanh(\mathbf{W}_n \times \mathbf{r}_i+\mathbf{b}_n),\\
\alpha^n_i & = \frac{\exp(a^n_i)}{\sum_{j=1}^N{\exp(a^n_j)}},
\end{split}
\end{equation}
where $\mathbf{q}_n$, $\mathbf{W}_n$ and $\mathbf{b}_n$ are the parameters in the news attention network.
The final representation of a user is the summation of the representations of news browsed by this user weighted by their attention weights: 
$\mathbf{u} = \sum_{i=1}^N{\alpha_i^n \mathbf{r}_i}$, where $N$ is the number of browsed news.

\subsection{Click Predictor}
The \textit{click predictor} module is used to predict the probability of a user browsing a candidate news based on their representations.
Denote the representation of a candidate news $D^c$ as $\mathbf{r}_c$ and the representation of user $u$ as $\mathbf{u}$.
Following~\cite{okura2017embedding}, the click probability score $\hat{y}$ is calculated by the inner product of the representation vectors of user $u$ and the candidate news $D^c$, i.e., $\hat{y} = \mathbf{u}^T\mathbf{r}_c.$
We also explored other probability computation methods such as multi-layer neural networks~\cite{wang2018dkn}.
We find that the inner product is not only the one with the best time efficiency but also the one with the best performance, which is consistent with~\cite{okura2017embedding}.

\subsection{Model Training}

Motivated by~\cite{huang2013learning} and~\cite{zhai2016deepintent}, we propose to use negative sampling techniques for model training.
For each news browsed by a user which is regarded as a positive sample, we randomly sample $K$ news which are presented in the same session but are not clicked by this user as negative samples.
We then jointly predict the click probability scores of the positive news $\hat{y}^+$ and the $K$ negative news $[\hat{y}_1^-, \hat{y}_2^-, ...,\hat{y}_K^-]$.
In this way, we formulate the news click prediction problem as a pseudo $K+1$-way classification task.
We normalize these click probability scores using softmax to compute the posterior click probability of a positive sample as follows:
\begin{equation}
p_i=\frac{\exp(\hat{y}_i^+)}{\exp(\hat{y}_i^+)+\sum_{j=1}^{K}{\exp(\hat{y}_{i,j}^-)}},
\end{equation}
where $\hat{y}_i^+$ is the click probability score of the $i$-th positive news, and $\hat{y}_{i,j}^-$ is the click probability score of the $j$-th negative news in the same session with the $i$-th positive news.
The loss function $\mathcal{L}$ in our approach for model training is the negative log-likelihood of all positive samples, which can be formulated as follows:
\begin{equation}
 \mathcal{L}=-\sum_{i\in \mathcal{S}}\log(p_i),
 \label{eq}
\end{equation}
where $\mathcal{S}$ is the set of the positive training samples.

\section{Experiments}\label{sec:Experiments}

\subsection{Datasets and Experimental Settings}

Our experiments were conducted on a real-world dataset, which was constructed by randomly sampling user logs from MSN News\footnote{https://www.msn.com/en-us/news} in one month, i.e., from December 13, 2018 to January 12, 2019.
The detailed statistics of the dataset are shown in Table~\ref{dataset}.
We use the logs in the last week as the test set, and the rest logs are used for training.
In addition, we randomly sampled 10\% of training samples for validation.

\begin{table}[h]
\resizebox{0.48\textwidth}{!}{
\begin{tabular}{|c|r|c|r|}
\hline
\textbf{\# users}       & 10,000     & \textbf{avg. \# words per title} & 11.29      \\ \hline
\textbf{\# news}        & 42,255     & \textbf{avg. \# words per body} & 730.72 \\ \hline
\textbf{\# impressions} & 445,230    & \textbf{\# positive samples}     & 489,644 \\ \hline
\textbf{\# samples}     & 7,141,584 & \textbf{\# negative samples}     & 6,651,940     \\ \hline
\end{tabular}
}
\vspace{-0.1in}
\caption{Statistics of our dataset.}\vspace{-0.15in}
\label{dataset}
\end{table}

In our experiments, the dimensions of word embeddings and category embeddings were set to 300 and 100 respectively.
We used the pre-trained Glove embedding~\cite{pennington2014glove} in our approach. 
The number of CNN filters was set to 400, and the window size was 3.
The dimension of the dense layers in the category view was set to 400.
The sizes of attention queries was set to 200.
The negative sampling ratio $K$ was set to 4.
We applied 20\% dropout ~\cite{srivastava2014dropout} to each layer in our approach to mitigate overfitting.
Adam~\cite{kingma2014adam} was used as the optimization algorithm.
The batch size was set to 100.
These hyperparameters were selected according to the validation set.
The metrics in our experiments include the average AUC, MRR, nDCG@5 and nDCG@10 scores over all impressions.
We independently repeated each experiment for 10 times and reported the average performance.

\subsection{Performance Evaluation}

First, we will evaluate the performance of our approach by comparing it with several baseline methods.
The methods to be compared include:
(1) \textit{LibFM}~\cite{rendle2012factorization}, which is a state-of-the-art feature-based matrix factorization method;
(2) \textit{CNN}~\cite{kim2014convolutional}, applying CNN to the concatenation of news titles, bodies and categories to learn news representations;
(3) DSSM~\cite{huang2013learning}, a deep structured semantic model with word hashing via character trigram and multiple dense layers;
(4) \textit{Wide\&Deep}~\cite{cheng2016wide}, using the combination of a wide linear channel and a deep neural network channel;
(5) \textit{DeepFM}~\cite{guo2017deepfm}, using a combination with factorization machines and neural networks;
(6) \textit{DFM}~\cite{lian2018towards}, a deep fusion model by using combinations of dense layers with different depths;
(7) \textit{DKN}~\cite{wang2018dkn}, a deep news recommendation method based on knowledge-aware CNN;
(8) \textit{NAML}, our neural news recommendation approach with attentive multi-view learning.
The experimental results of different methods are summarized in Table~\ref{table.result}.

According to Table~\ref{table.result}, We have several observations.
First, the methods based on neural networks (e.g., \textit{CNN}, \textit{DSSM} and \textit{NAML}) outperform traditional matrix factorization methods such as \textit{LibFM}.
This is probably because neural networks can learn better news and user representations  than traditional matrix factorization methods.

Second, the methods using negative sampling (\textit{DSSM} and \textit{NAML}) outperform the methods without negative sampling (e.g., \textit{CNN}, \textit{DFM} and \textit{DKN}).
This is probably because the methods without negative sampling are trained on a balanced dataset, which is not consistent with real-word news recommendation scenarios.
\textit{DSSM} and our \textit{NAML} approach can utilize the information from much more negative samples than other baseline methods, which is useful for learning more accurate news representations.

Third, our approach can consistently outperform other baseline methods.
Since the characteristics of news titles, bodies and categories are usually divergent, simply concatenating the features extracted from these sources may be sub-optimal.
Different from baseline methods, our approach uses a multi-view framework to incorporate different information as different views.
In addition, different words, news and views may have different informativeness for representing news and users.
Our approach can simultaneously select important words, news and views, which can build more informative news and user representations.

\begin{table}[!t]
	\centering
	\resizebox{0.48\textwidth}{!}{
\begin{tabular}{|c|c|c|c|c|}
\hline
    \textbf{Methods}         & \textbf{AUC}             & \textbf{MRR}             & \textbf{nDCG@5}          & \textbf{nDCG@10}         \\ \hline
LibFM        & 0.5880          & 0.3054          & 0.3202          & 0.4090          \\
CNN          & 0.5909          & 0.3068          & 0.3221          & 0.4109         \\
DSSM         & 0.6114          & 0.3188          & 0.3261          & 0.4185          \\ 
Wide\&Deep & 0.5846          & 0.3009          & 0.3167          & 0.4062          \\
DeepFM       & 0.5896          & 0.3066          & 0.3221          & 0.4117          \\ 
DFM          & 0.5996          & 0.3133          & 0.3288          & 0.4165          \\
DKN          & 0.5966          & 0.3113          & 0.3286          & 0.4165          \\ \hline
NAML*          & \textbf{0.6434} & \textbf{0.3411} & \textbf{0.3670} & \textbf{0.4501} \\ \hline
\end{tabular}
}
	\caption{The results of different methods. *The improvement over all baseline methods is significant at the level $p<0.001$.}\label{table.result}\vspace{-0.1in}
\end{table}
\subsection{Effectiveness of Attentive Multi-View Learning}

\begin{figure}[t]
	\centering
	\subfigure[Multi-view learning.]
	{
		\label{fig.view} 
		\includegraphics[width=0.22\textwidth]{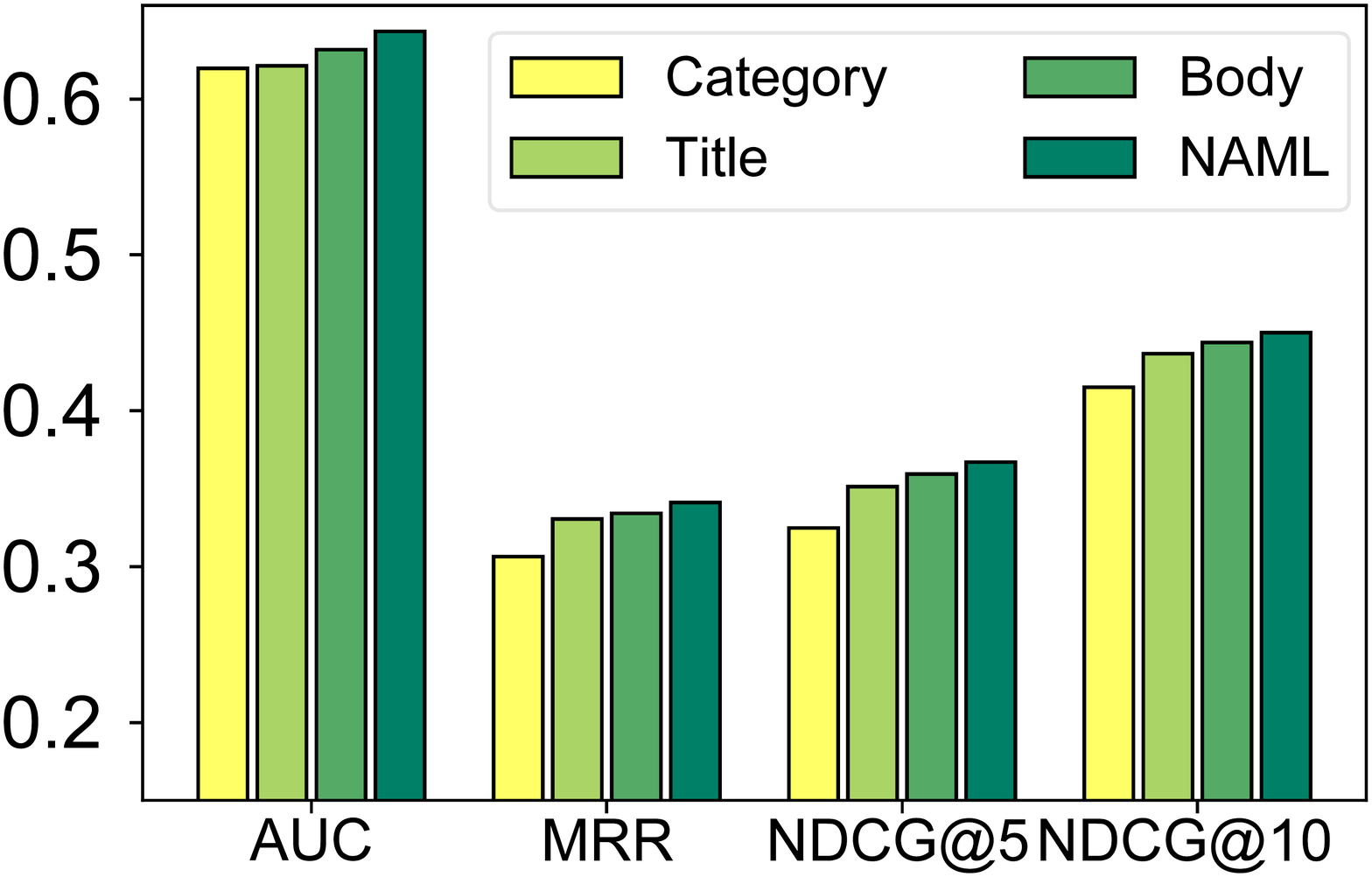}
	}
	\subfigure[Attention networks.]
	{
		\label{fig.att} 
		\includegraphics[width=0.22\textwidth]{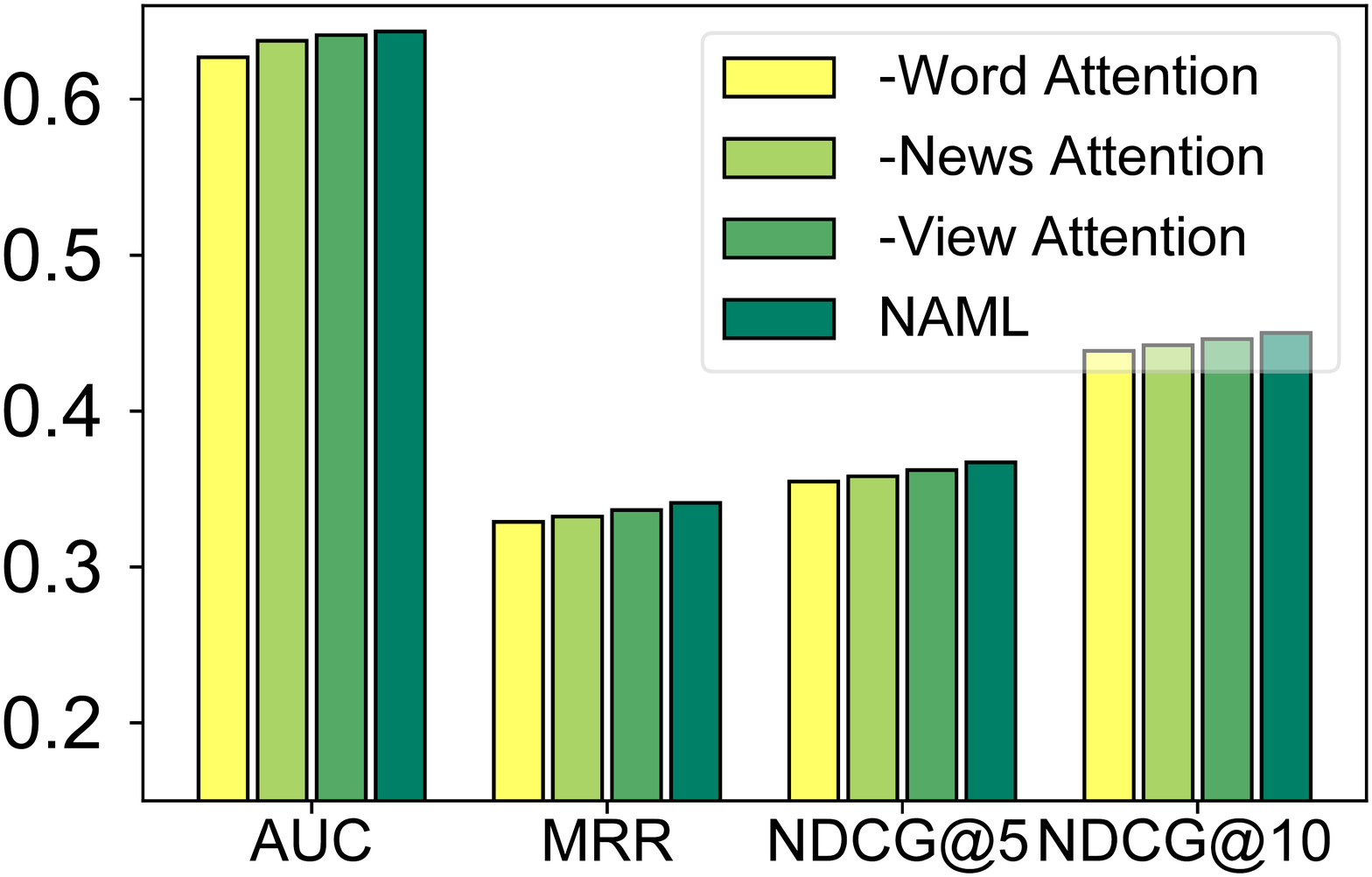}
	}
	\caption{The effectiveness of the multi-view learning framework and the attention networks in our approach.}\label{fig.vatt}
	\vspace{-0.1in}
\end{figure}

In this section, we conducted several experiments to validate the effectiveness of attentive multi-view learning framework in our model.
First, we explore the effectiveness of  multi-view framework in our \textit{NAML} approach.
The performance of \textit{NAML} and its variants with different combinations of views is shown in Fig.~\ref{fig.view}.
According to Fig.~\ref{fig.view}, we have several observations.
First, the model with the body view achieves better performance than those with the title or category view only.
This is intuitive because the bodies of news usually contain the original information of news, and can provide rich information for modeling news topics.
Second, the title and category views are also informative for news recommendation. 
This is probably because titles usually have decisive influence on users' reading behaviors.
Thus, incorporating news titles is useful for modeling the characteristics of news and users.
In addition, since categories of news are important clues of news topics, incorporating the category view is also useful for recommendation.
Third, combine all three views can further improve the performance of our approach.
These results validate the effectiveness of our multi-view framework.

Next, we conducted experiments to validate the effectiveness of the attention mechanism at word-, news- and view-level.
The performance of \textit{NAML} and its variant with different combinations of attention networks is shown in Fig.~\ref{fig.att}.
According to Fig.~\ref{fig.att}, we have several observations.
First, the word-level attention network can effectively improve the performance of our approach.
This is probably because words are basic units in titles and bodies to convey their meanings, and different words usually have different informativeness for learning news representations.
Our approach can recognize and highlight the important words, which is useful for learning more informative news representations.
Second, the news-level attention is also important to improve the performance of our approach.
This is probably because news browsed by the same user usually have different informativeness for learning the representations of this user, and recognizing  important news is useful for learning high-quality user representations.
Third, using view-level attention network can also improve the performance of our approach.
Since the three views can also have different informativeness for modeling news and users, evaluating the importance of views may be useful for improving the performance of our approach.
Fourth, combining all attention networks can further improve the performance of our approach.
These results validate the effectiveness of attention mechanism in our approach.

\subsection{Visualization of Attention Weights}
\begin{figure}[t]
	\centering
	\subfigure[The mean and standard deviation of attention weights of views.]
	{
		\label{fig.viewatt0} 
		\includegraphics[width=0.24\textwidth]{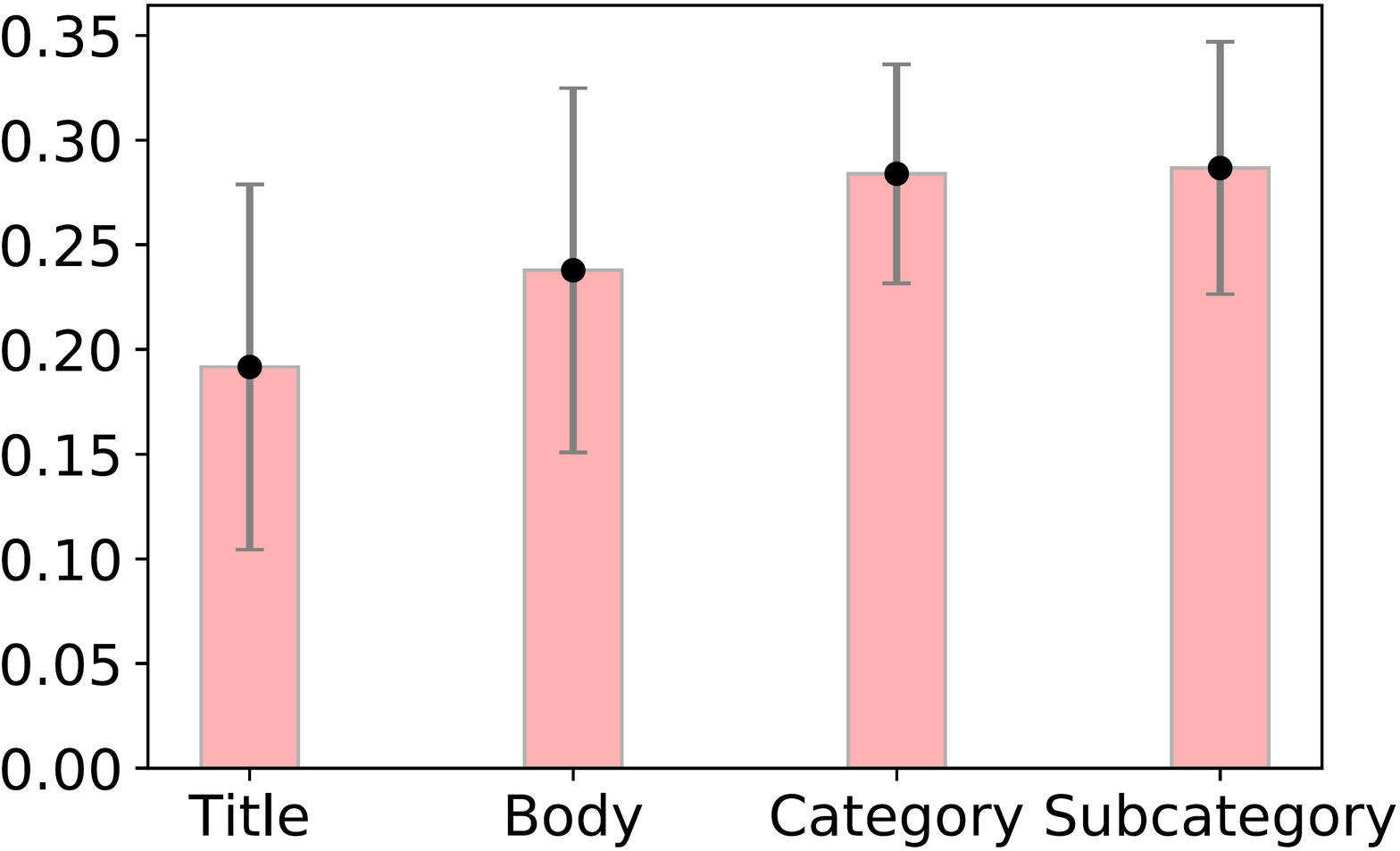}
	}
	\subfigure[Distributions of attention weights of views.]
	{
		\label{fig.viewatt} 
		\includegraphics[width=0.21\textwidth]{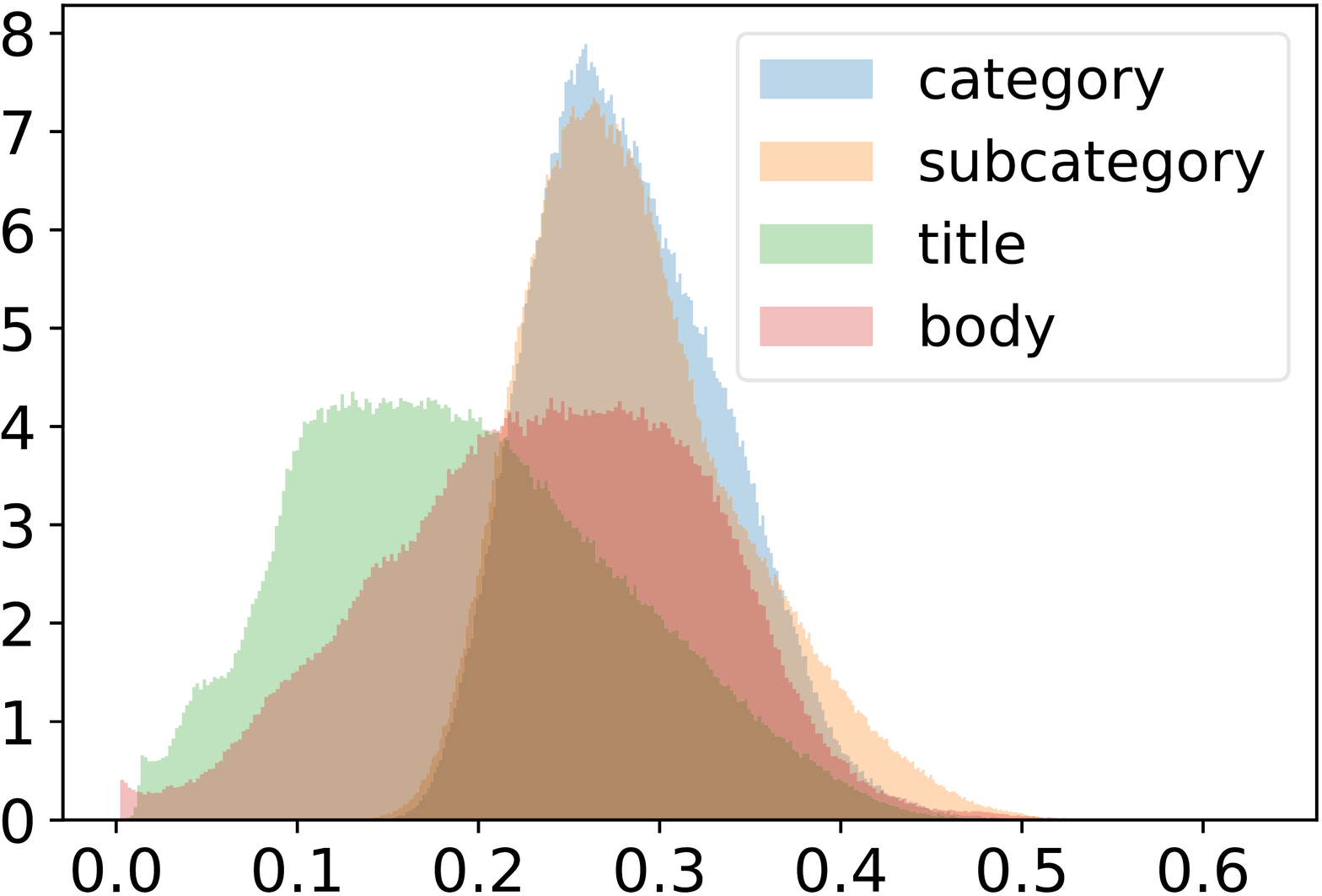}
	}	
	\caption{Visualization of the view-level attention weights.}\label{fig.vatt}
	\vspace{-0.1in}
  \end{figure}
In this section, we will visually explore the effectiveness of the attention mechanism in our approach.
First, we want to visualize the view-level attention weights of the three views.
The results are shown in Fig.~\ref{fig.vatt}.
From Fig.~\ref{fig.viewatt0}, we find the average attention weights of the body view are higher than those of the title view.
This may be because bodies usually convey the original meanings of news and contain richer  information than titles.
Thus, the attention weights of the body view is higher.
Second, we find it is interesting that the category view gains the highest attention weights.
This may be because categories are very important clues for inferring news topics, which are very informative for learning accurate news representations.
The visualization results of distributions of the view-level attention weights are shown in Fig.~\ref{fig.viewatt}.
We find that the attention weights on the title and body views are small for many samples.
This may be because some news titles and bodies are vague and uninformative for learning news representations.
In these cases, categories are can provide rich complementary information for recommendation, which validate the effectiveness of our multi-view framework.

\begin{figure}[t]
	\centering
		\includegraphics[width=0.48\textwidth]{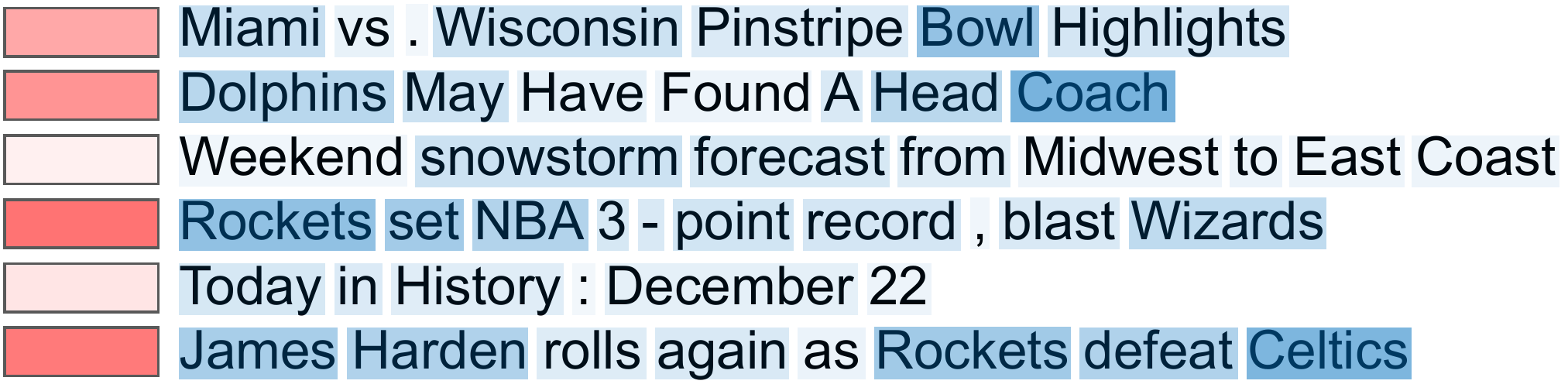}
	\caption{Visualization of the word- and news-level attention weights. Darker colors represent higher attention weights.}\label{fig.newsatt}	\vspace{-0.1in}
  \end{figure}
Then, we want to visually explore the effectiveness of the word- and news-level  attention networks.
For the word-level attention network, we only visualize the attention weights of the words in news titles since news bodies are usually too long.
The visualization results of several clicked  news from a randomly selected users are shown in Fig.~\ref{fig.newsatt}.
From Fig.~\ref{fig.newsatt}, we find the attention network can effectively recognize important words within news articles.
For example, the word ``NBA'' and ``Rockets'' are highlighted, since these words are very informative for representing news topics, while the word ``December'' is assigned a low attention weight since it is less informative for learning news topics.
These results show that our approach can learn informative news representations by recognizing important words.
From Fig.~\ref{fig.newsatt}, we also find our approach can also effectively recognize important news of a user.
For example, the news ``Rockets set NBA 3 - point record , blast Wizards'' gains high attention weight because it is very informative for modeling user preferences, while the news ``Today in History : December 22'' gains low attention weight, since it is not very informative for representing users.
These results show that our model can effectively evaluate the different importance of words and news articles for learning informative news representations.

\section{Conclusion}\label{sec:Conclusion}

In this paper, we propose a neural news recommendation approach with attentive multi-view learning (NAML).
The core of our approach is a news encoder and a user encoder.
In the news encoder, we propose a multi-view learning framework to learn unified news representations by incorporating titles, bodies and categories as different views of news.
We also apply attention mechanism to news encoder to select important words and views for learning informative news representations.
In the user encoder, we learn representations of users from their browsed news, and apply a news attention network to select important news for learning informative user representations.
Extensive experiments on real-world dataset show our approach can improve the performance of news recommendation and outperform many baseline methods.
\section*{Acknowledgments}
The authors would like to thank Microsoft News for providing technical support and data in the experiments, and Jiun-Hung Chen (Microsoft News) and Ying Qiao (Microsoft News) for their support and discussions.
This work was supported by the National Key Research and Development Program of China under Grant number 2018YFC1604002, the National Natural Science Foundation of China under Grant numbers U1836204, U1705261, U1636113, U1536201, and U1536207, and the Tsinghua University Initiative Scientific Research Program under Grant number 20181080368.
\bibliographystyle{named}
\bibliography{ijcai19}

\end{document}